\documentclass[10pt,letterpaper]{article}

\usepackage{cogsci}

\cogscifinalcopy % Uncomment this line for the final submission 
\usepackage{graphicx}
\usepackage{pslatex}
\usepackage{apacite}
\usepackage[font=small]{caption}
\usepackage{float} % Roger Levy added this and changed figure/table
\title{A Developmentally-Inspired Examination of Shape versus Texture Bias in Machines}
 
\author{{\large \bf Alexa R. Tartaglini, Wai Keen Vong, and Brenden M. Lake} \\
\{art481, waikeen.vong, brenden\}@nyu.edu\\
New York University}

\begin{document}

\maketitle

\begin{abstract}
Early in development, children learn to extend novel category labels to objects with the same shape, a phenomenon known as the shape bias. Inspired by these findings, \citeA{geirhos2019imagenet} examined whether deep neural networks show a shape or texture bias by constructing images with conflicting shape and texture cues. They found that convolutional neural networks strongly preferred to classify familiar objects based on texture as opposed to shape, suggesting a texture bias. However, there are a number of differences between how the networks were tested in this study versus how children are typically tested. In this work, we re-examine the inductive biases of neural networks by adapting the stimuli and procedure from \citeA{geirhos2019imagenet} to more closely follow the developmental paradigm and test on a wide range of pre-trained neural networks. Across three experiments, we find that deep neural networks exhibit a preference for shape rather than texture when tested under conditions that more closely replicate the developmental procedure. % We also discover a setting of novel stimuli which do not induce a shape or texture bias in a randomly-initialized model, indicating that test stimuli can be “tuned” to achieve a neutral shape bias test and help distill the role of learning in shape-based generalization.


\end{abstract}

\section{Introduction}

\begin{figure}
    \centering
    \includegraphics[width=0.48\textwidth]{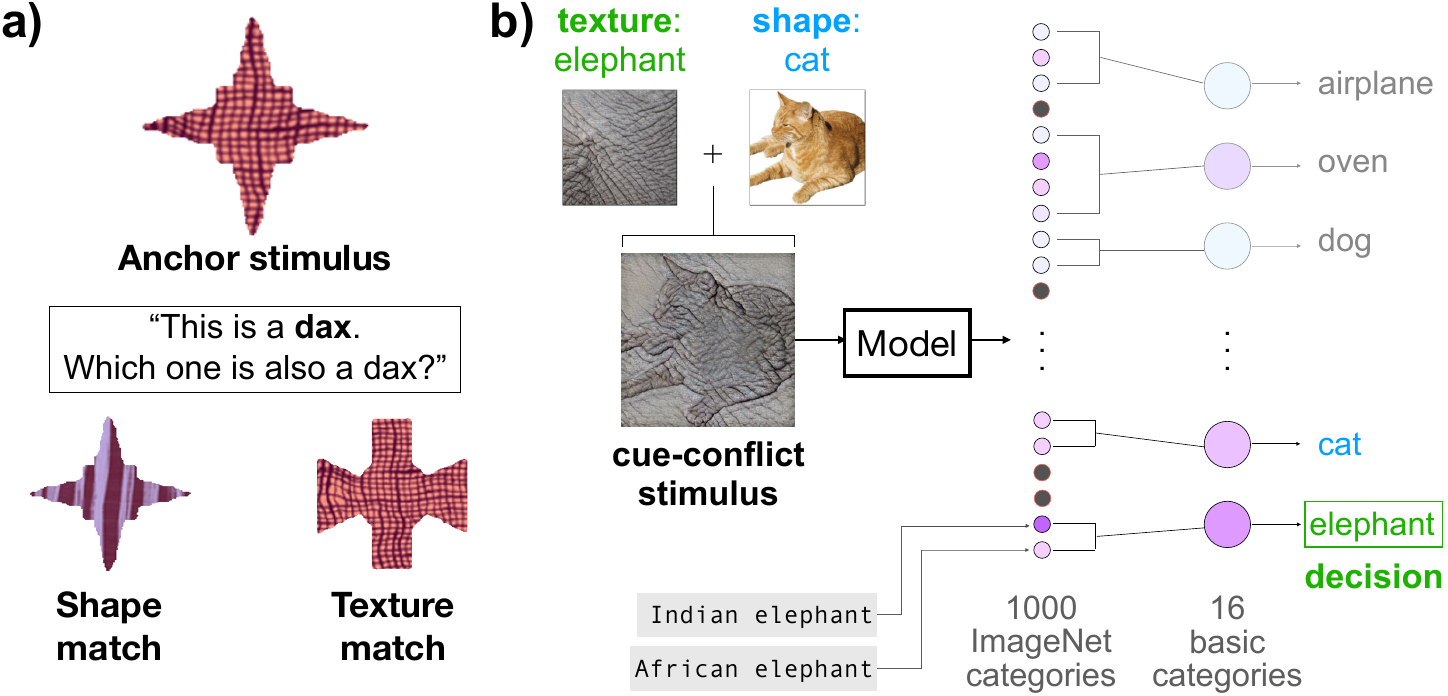}
    \caption{\textbf{Two different approaches to measuring shape bias}. In the standard developmental procedure (left), an anchor stimulus (top) is presented and assigned a novel label. The participant is then asked whether a shape match (bottom left) or texture match (bottom right) has the same novel label as the anchor stimulus, and shape bias is calculated as the proportion of participants who prefer the shape match. In the shape bias procedure from \citeA{geirhos2019imagenet} (right), a cue-conflict stimulus with a cat shape and elephant texture is presented to a model, and a subset of the 1000 resulting ImageNet class probabilities (small circles) are averaged to attain scores for 16 basic categories. For example, probabilities for the ImageNet categories ``Indian elephant'' and ``African elephant'' are averaged to obtain a score for the basic category ``elephant''. All other ImageNet probabilities (dark gray circles) are ignored. The model's decision is considered shape biased if it predicts the shape category (cat), texture biased if it predicts the texture category (elephant), or is discarded if it predicts any other category.}
    \label{fig:procedure}
\end{figure}

When presented with a new object and its label (e.g. ``dax''), how do humans determine how to generalize the label to other objects? Starting around the age of two, children preferentially generalize novel category labels to solid objects of the same shape rather than the same size, color, or texture, a phenomenon known as the \textbf{shape bias} \cite{landau1988importance}. The standard experimental approach to measuring the shape bias in developmental psychology involves first presenting the participant with an anchor stimulus consisting of a novel shape and texture (see Figure~\ref{fig:procedure}(a) for an illustration). This is followed by the presentation of additional stimuli that match the anchor stimulus in either shape, texture, size, or color (but not in any other dimension). Participants are asked which of the additional stimuli are the same category as the anchor; choosing the stimulus that matches along the shape dimension at levels above chance indicates a shape bias. The emergence of the shape bias indicates a crucial developmental shift in which children begin to recognize that shape information is a reliable indicator of object names, facilitating the acceleration of noun learning \cite{smith2002object, diesendruck2003specific, gershkoff2004shape}. The shape bias strengthens as we grow older and gain more visual and linguistic experience; in fact, by the time we are adults, shape appears to serve as the primary predictor of our recognition of familiar categories \cite{biederman1995visual}. 

Due to its importance in understanding generalization and learning in humans, multiple computational accounts of the shape bias have been proposed. These include models based on associative learning \cite{samuelson2002statistical, colunga2005lexicon} or overhypotheses in hierarchical Bayesian models \cite{kemp2007learning}, although one limitation is that these models used simplified feature representations for capturing shape rather than naturalistic images. In the past decade, following the impressive achievements of deep neural networks (DNNs) on tasks such as image classification \cite{krizhevsky2012imagenet}, there has been a push to better understand how machines encode and process images and whether they can be used as computational accounts of human vision \cite{yamins2014performance, geirhos2021partial}. In particular, this has led to a renewed interest in the shape bias, with initial accounts demonstrating that DNNs did have a preference for shape. \citeA{ritter2017cognitive} showed that pre-trained convolutional neural networks (CNNs) preferred to categorize novel objects on the basis of shape rather than color, and \citeA{feinman2018learning} found that simple neural networks could learn a shape bias from very few examples. 

However, while this earlier work showed that DNNs displayed a shape bias, this may have been the result of limiting the comparison to shape versus color. More recent work has argued that deep neural networks actually rely on local texture information rather than global shape information for classification \cite{baker2018deep, brendel2019approximating, geirhos2019imagenet}. In particular, \citeA{geirhos2019imagenet} conducted a deeper examination by measuring shape versus texture bias in ImageNet pre-trained deep neural networks. They created a set of novel cue-conflict stimuli by overlaying the shape information of one category with the texture information of another category via style transfer \cite{gatys2016image}. Shape bias was then calculated as the proportion of trials in which a model classified a cue-conflict stimulus as its shape category divided by the total number of classifications to either its shape or texture category (and similarly for texture bias; see Figure~\ref{fig:procedure}(b) for procedure details). Using this procedure, they found that ImageNet pre-trained CNNs were actually strongly biased towards texture, while humans tested using the same stimuli showed a strong preference for shape, highlighting a large discrepancy between human and machine behavior. In light of this finding, additional research has aimed to find ways of reducing the degree of texture bias in deep neural networks, such as work showing that shape bias can be increased via different kinds of data augmentations \cite{hermann2020origins}, or that vision transformers show less of a texture bias than CNNs \cite{tuli2021convolutional}.

\begin{figure}
    \centering
    \includegraphics[width=0.48\textwidth]{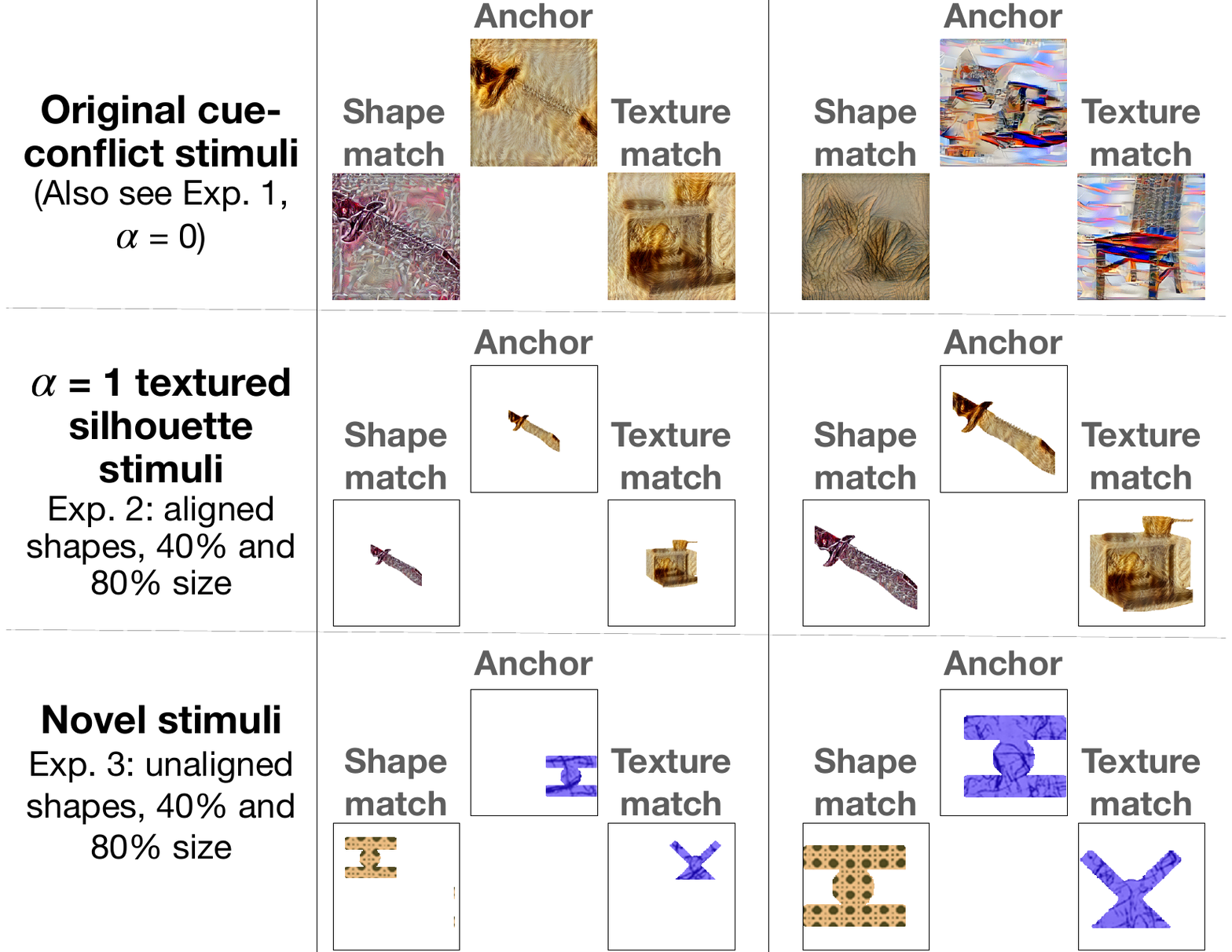}
    \caption{\textbf{Example triplets from several experimental conditions}. Each row consists of two triplet-based trials, each with an anchor stimulus, a shape match, and a texture match. The first row demonstrates two triplets created from the original cue-conflict stimuli. For the second and third rows, the triplets on the left demonstrate the 40\% size condition, while the triplets on the right demonstrate the 80\% size condition. The textured silhouette stimuli in the second row also demonstrate the aligned shape condition, while the novel stimuli in the third row demonstrate the unaligned shape condition.}
    \label{fig:stimuli}
\end{figure}

Despite the significant attention this finding has received, there are a number of important differences between the procedure used by \citeA{geirhos2019imagenet} and the procedure used in developmental psychology for assessing the shape bias, making it difficult to directly compare these recent findings with traditional findings. First, as shown in Figure~\ref{fig:procedure}, the style transfer method for creating the cue-conflict stimuli from \citeA{geirhos2019imagenet} produces stimuli where both the texture information may be inadvertently emphasized and the shape information de-emphasized, by covering the entire image---both the shape itself and the background---with the texture pattern. On the other hand, in developmental studies, the stimulus texture is always contained within the shape boundary, more closely reflecting how texture applies to objects in the real world; in fact, physical objects are often used rather than images of stimuli on a computer screen. Second, the procedure is limited to evaluating shape bias for 16 highly \textit{familiar} categories (due to the use of a model's output layer from the set of 1000 ImageNet categories), whereas developmental studies focus on testing shape bias via \textit{novel} shapes. Finally, the use of model classification outputs to determine a shape or texture decision for each individual stimulus differs from the typical procedure showing an anchor against shape or texture matches. In fact, the procedure from \citeA{geirhos2019imagenet} requires discarding a large proportion of trials that do not result in a correct shape or texture classification in their measure of shape bias.

Motivated by these differences, we re-examine the texture bias claim and outline an alternative procedure for measuring shape bias in artificial neural networks that is more closely aligned with the developmental procedure. In Experiment 1, we adapt the cue-conflict stimuli from \citeA{geirhos2019imagenet} to parametrically reduce the saliency of the background texture. Our main result shows that across all models tested, removing the influence of background texture results in a preference for shape over texture. In Experiment 2, we conduct additional tests to check the robustness of this finding by varying the size and positioning of the adapted stimuli, showing that this preference for shape over texture holds across different presentation conditions. Finally, in Experiment 3, we test the shape bias using a set of stimuli with novel shapes and textures, which most closely replicates the developmental procedure. We find that the overall degree of shape bias across models is lower for the novel stimuli compared to the adapted stimuli from Experiment 2. Furthermore, pre-trained models exhibit a higher rate of shape-based responses than their untrained counterparts, indicating that training increases shape bias. The shape bias testing code and datasets are available at: \url{https://github.com/alexatartaglini/developmental\linebreak -shape-bias}.

\section{Methods}

\textbf{Measuring Shape Bias.} 
In this section, we describe the procedure used in the following three experiments to measure shape bias in the models. Following the developmental procedure from \citeA{landau1988importance}, a trial consists of a triplet of image stimuli: an anchor stimulus with a given shape and texture, a shape match stimulus that shares only its shape with the anchor, and a texture match stimulus that shares only its texture with the anchor. We assembled a large number of unique triplets for a given dataset by first considering each stimulus in the dataset as an anchor, then selected shape and texture matches for the anchor from other stimuli that have the same shape or texture class respectively.\footnote{Only shapes and textures from the same source image are considered to be a match; for example, if two stimuli both have the shape of a cat but use two different pictures of cats, they are not considered shape matches, and similarly for texture matches.}

To process a trial, all three images in a triplet were passed individually to a given pre-trained model up to the penultimate layer to extract three embeddings of visual features. We then determined whether the model considered the shape or texture match to be more similar to the anchor by computing the cosine similarity between the anchor and the two matches.\footnote{We also ran all experiments using the dot product and Euclidean distance as alternative measures of perceptual similarity. We found that the results were very similar across all distance metrics used, so only results using cosine similarity are reported.} If the cosine similarity between the anchor and shape match is higher, the trial is considered a shape decision. Otherwise, it is considered a texture decision. The resulting shape bias of a model is computed as the proportion of the number of its shape decisions to the total number of trials. Because no trials are discarded with this method, texture bias of the models equals the remaining number of texture decisions over the total number of trials.  

\textbf{Models.}
We measured shape bias for a variety of models with different architectures, types of supervision, and training data. This included a \textbf{ResNet-50} convolutional network \cite{he2016deep} and a \textbf{ViT-B/16} vision transformer \cite{dosovitskiy2021image}. As a baseline, we first ran all experiments with 10 randomly-initialized ResNet-50s and ViT-B/16s then computed the average shape bias for each across all 10 models. These models are referred to as \textbf{random ResNet-50} and \textbf{random ViT-B/16} respectively. We also tested supervised variants of ResNet-50 and ViT-B/16 that were pre-trained on ImageNet \cite{deng2009imagenet}, as well as a self-supervised ResNet-50 trained via \textbf{DINO} \cite{caron2021emerging}, that did not require labels during pre-training.

We also included \textbf{CLIP ViT-B/16} \cite{radford2021learning}, which was pre-trained on a dataset of 400 million image-caption pairs via contrastive learning, and was recently been shown to be most comparable to human vision on a range of benchmarks \cite{geirhos2021partial}. Finally, we tested \textbf{SAYCam-S} model \cite{orhan2020self}, which uses a different convolutional neural network (ResNeXt-50, \shortciteNP{xie2017aggregated}) and was trained using a self-supervised objective on a longitudinal egocentric dataset of headcam footage filmed from the perspective of one child sampled regularly from the age of 6 months to 32 months. We included this final model because the footage is recorded during the developmental period and in the typical environment that English-speaking children tend to acquire a shape bias (although this model does not receive the kind of labeled supervision thought to be important in acquiring the shape bias, \citeNP{smith2002object, gershkoff2004shape}). 

% \begin{figure}
%     \centering
%     \includegraphics[width=0.48\textwidth]{figures/conditions.pdf}
%     \caption{\textbf{Examples of stimuli with varying conditions used in Experiments 1-3}. \textbf{Top row}: Increasing $\alpha$ decreases background texture salience. When $\alpha=0$ (leftmost stimulus), the stimuli are equivalent to the original stimuli provided by \citeA{geirhos2019imagenet}. \textbf{Middle row}: Starting with the $\alpha=1$ textured silhouette stimuli, we vary the size and alignments of the shapes. The aligned shape condition is displayed. \textbf{Bottom row}: We vary size and alignment for the novel stimuli. Displayed is the unaligned shape condition; the largest shapes are very nearly but not exactly aligned when randomly positioned.}
%     \label{fig:conditions}
% \end{figure}

\section{Experiment 1}
In this experiment, we examined the possibility that the texture bias observed in convolutional neural networks may be due to an over-emphasis of texture in the cue-conflict stimuli used in \citeA{geirhos2019imagenet}. Specifically, the texture information in these test stimuli is highly salient, covering both the shape and background and obscuring some of the underlying shape information (see the top row of Figure~\ref{fig:stimuli}). This is in contrast to the stimuli used in the developmental setup, in which the texture is contained within the shape of a stimulus and presented on a white or neutral background. 

\begin{figure}
    \centering
    \includegraphics[width=0.48\textwidth]{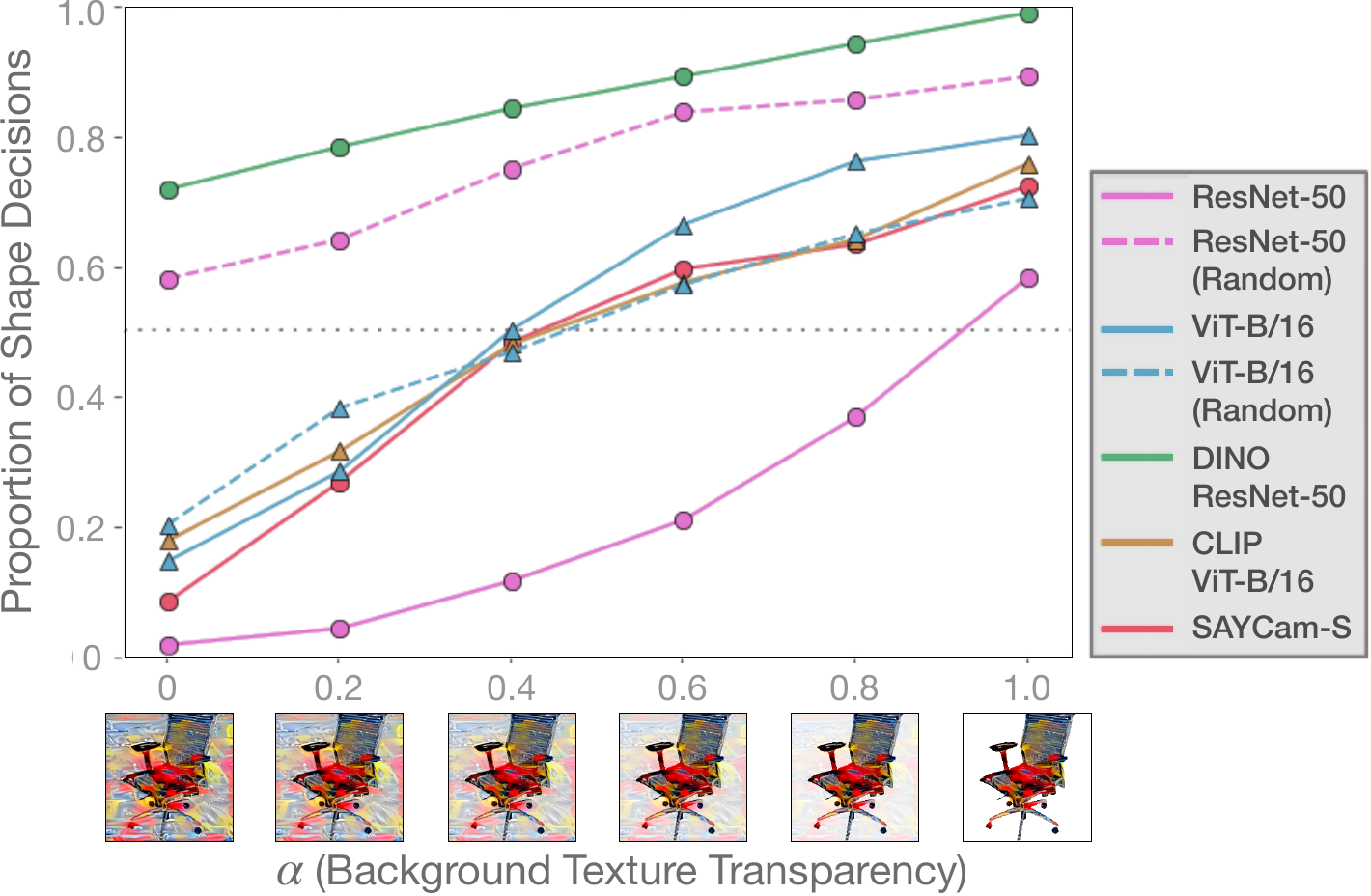}
    \caption{\textbf{Shape bias vs. $\alpha$}. Under each $\alpha$ value on the x-axis, there is an example of a stimulus demonstrating the degree of background transparency. Each line represents a different model (see the legend to the right); CNN architectures are represented with circular markers, while vision transformers are represented with triangular markers. Results for randomly-initialized models are plotted with a dashed line. The horizontal, dotted gray line represents chance levels of shape bias; any model above this line is shape biased, while any model below is texture biased. All models show an increase in shape bias as the texture in the background becomes more transparent.}
    \label{fig:alpha}
\end{figure}

\textbf{Dataset.}
In order to bridge these visual differences, we modified the original cue-conflict stimuli created by \citeA{geirhos2019imagenet} by parametrically decreasing the opacity of the background texture in varying degrees. This was achieved by making use of the separate ``filled silhouette'' dataset \cite{geirhos2019imagenet}, which contained a silhouette of each shape instance used in the original cue-conflict stimuli. The silhouette of each shape instance was used to create a mask of the background for each matching cue-conflict stimulus. These masks were then superimposed onto the corresponding cue-conflict stimuli, obscuring the texture background with white pixels and highlighting the shape of the stimulus. We refer to these modified stimuli as the ``textured silhouette'' stimuli. The opacity of the white background mask was controlled by a variable $\alpha \in [0, 1]$, where $\alpha = 0$ is equivalent to the original cue-conflict stimuli, and $\alpha = 1$ removes all background texture and replaces it with a white background. We created distinct datasets for $6$ equally spaced $\alpha$ values, as shown in Figure~\ref{fig:alpha}.

All possible (anchor, shape match, texture match) triplets of stimuli were generated as described in the \textbf{Measuring Shape Bias} section for each $\alpha$-valued textured silhouette dataset. In the original dataset used by \citeA{geirhos2019imagenet}, stimuli with certain shapes and textures appeared more frequently than others, which resulted in a larger number of possible triplets with these anchor stimuli. To ensure that all shape and texture classes were equally represented in the final shape bias computation, we randomly selected $28$ unique triplets for each of the $1,200$ anchor stimuli, producing a total of $33,600$ triplets used for evaluation. We repeated this procedure a total of 3 times, reporting the average measurements across replications.

% three times with replacement. Each of the three triplet selections consist of $33,600$ triplets. We compute shape bias three times, once for each of the three selections. We then report model shape bias as the average of these three measurements.

\textbf{Results and Discussion.}
When $\alpha = 0$ and the background texture is fully salient, we predictably find a pattern of results that mirrors the observations made by \citeA{geirhos2019imagenet}. Across the board, the models are highly texture biased, with the notable exception of the random ResNet-50 and DINO ResNet-50, which are weakly shape biased (see Figure~\ref{fig:alpha}). Surprisingly, we observed an even stronger texture bias than \citeA{geirhos2019imagenet}; for example, they observed about a 20\% shape bias in a pre-trained ResNet-50 compared to the mere 2\% shape bias we observed using the triplet-based procedure. As $\alpha$ increases and background texture salience decreases, all models displayed a monotonic increase in shape bias, although at varying rates. The vision transformer models including the CLIP model and the SAYCam-S model showed a substantial increase in shape bias, and were between 70-80\% shape biased when the background is completely white. On the other hand, the supervised ResNet-50 remained relatively texture biased throughout, only attaining a slight shape bias when $\alpha = 1$. Still, this increase is surprising given the extremely strong texture bias observed for ResNet-50 when $\alpha=0$. DINO ResNet-50 is the most shape biased for all $\alpha$ values and attained nearly 100\% shape bias when the background texture is completely removed. This may be the result of the data augmentations used when training DINO such as Gaussian blur and color augmentation, which have been shown to increase shape bias in CNNs \cite{hermann2020origins}. Finally, random ResNet-50 also displayed a rather strong shape bias with low variance across seeds throughout, which is curious given its lack of training.\footnote{We also ran Experiment 1 for all supervised models using the original classification decision procedure from \citeA{geirhos2019imagenet} and found a very similar pattern of results.}

These results demonstrate that the degree of shape bias using the cue-conflict stimuli from \citeA{geirhos2019imagenet} is variable across architectures and can be modulated by altering the background of the stimuli. In the developmental test, the background is irrelevant, so shape bias as measured in artificial neural networks should also reflect this in their set-up. However, there may be alternative explanations for the pattern of results we observed in this experiment. One possibility is that the higher degree of shape bias is an artifact of the perfect alignment in position and size of the shape information between the anchor stimulus and shape match stimulus. In particular, this could explain the high degree of shape bias for the randomly-initialized ResNet-50, which may be especially sensitive to pixel overlap. 

\section{Experiment 2}
In this experiment, we expand upon the results from Experiment 1 and examined whether the high level of shape bias observed in all models when $\alpha = 1$ is robust to the size and position of the shape information in the stimuli. Real world objects vary in these factors, and thus, it is useful to understand how these properties influence model behavior. % We vary the size and positional alignment of the $\alpha = 1$ textured silhouette stimuli and measure shape bias for each of the resulting conditions. 

\begin{figure*}
    \centering
    \includegraphics[width=0.9\textwidth]{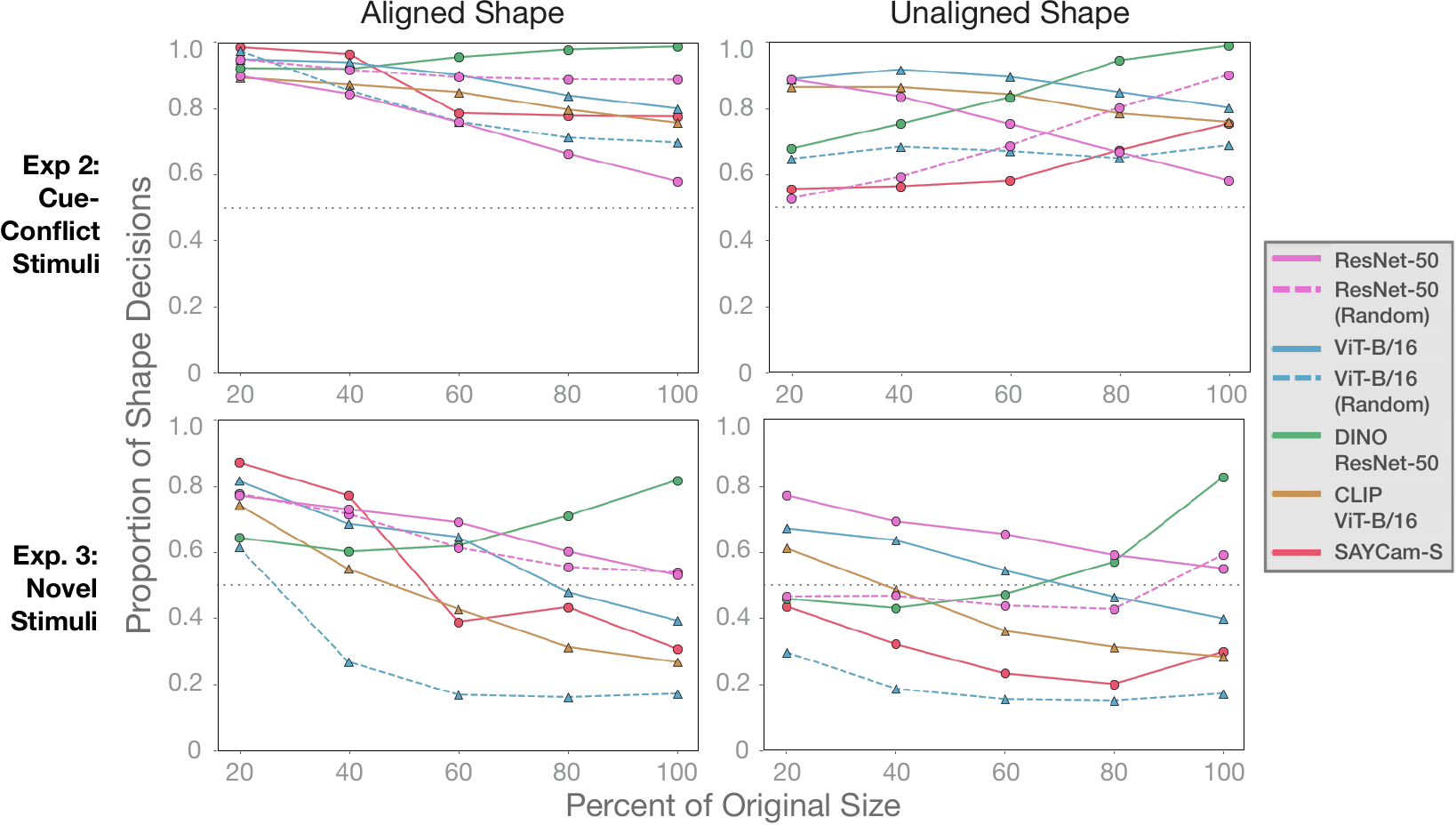}
    \caption{\textbf{Results from Experiments 2 and 3}. The rows indicate the experiment, while the columns represent the shape alignment condition with aligned shape results on the left and unaligned results on the right. \textbf{Top row}: using textured silhouette stimuli in Experiment 2, we find that all models are shape biased but respond differently to variations in shape size and alignment. \textbf{Bottom row}: using novel stimuli in Experiment 3, we find that all models are less shape-biased overall than Experiment 2. Unlike most of the models, DINO ResNet-50 increases in shape bias as shape size increases; also, for the unaligned shape condition, the random ResNet-50 hovers around chance levels of shape bias for all stimulus sizes with low variance.}
    \label{fig:align}
\end{figure*}

\textbf{Dataset.} Starting with the subset of the textured silhouette stimuli with a white background ($\alpha=1$), we created five different stimulus \textbf{size} conditions ranging from 20\% to 100\% of their original size, with each triplet being uniform in size. We also created two positional \textbf{alignment} conditions. In the aligned shape condition, all stimuli occupied the center of an image, resulting in perfect overlap between the anchor and shape match stimuli. In the unaligned shape condition, all stimuli were individually placed in random locations of the image; thus, in a given triplet, the three stimuli occupy different parts of the image. Figure \ref{fig:stimuli} provides examples of both the size and alignment variations.\footnote{Note that the 100\% size condition in Experiment 2 is technically always aligned and is equivalent to the $\alpha = 1$ textured silhouette stimuli from Experiment 1, so the rightmost data points of Figure~\ref{fig:alpha} are the same as the rightmost data points of the top two plots in Figure~\ref{fig:align}.} 
 
\textbf{Results and Discussion.}
When the shapes of the stimuli are aligned, all models were highly shape biased across all stimulus sizes, although they were relatively more shape biased for smaller stimuli (Figure~\ref{fig:align}, top left). When the shapes were 20\% of their original size, all models demonstrated a shape bias between 90\% and 100\% percent; for the largest stimuli, shape bias for the models ranged between 60\% and 100\%, with the ImageNet-trained ResNet-50 demonstrating the most significant decrease in shape bias. One possibility for this decrease is that shape bias decreased for larger shapes because a greater surface area allowed for a larger, more detailed patch of texture to be visible, thereby increasing the salience of the texture information. Another possibility is that the receptive field size or patch size in different models is smaller relative to the larger size conditions. When the shapes were randomly positioned in the image and no longer overlap in the shape unaligned condition, the models displayed more variation in their degree of shape bias but were still above the chance level for all stimulus sizes (Figure~\ref{fig:align}, top right). 
%ResNet-50, ViT-B/16, and CLIP ViT-B/16 all decrease in shape bias as the stimuli become larger, which may be explained by the fact that larger shapes allow for more texture information to be visible. However, the DINO ResNet-50, random ResNet-50, and SAYCam-S models all actually increase in shape bias as the stimuli become larger; this could be because these models prefer to attend to the small pieces of semantic information from the shape class, which become larger and more detailed as the shapes increase in size. Random ViT-B/16 hovers at around 65\% shape bias throughout. 

These results show that a robust shape bias can be observed in all models by removing the background texture from the stimuli used by \citeA{geirhos2019imagenet} despite the different variations in size and positioning we explored in this simulation. This strengthens the findings in Experiment 1 by ruling out the hypothesis that pixel-overlap between the anchor and shape match are driving the preference for shape; rather, the sensitivity to shape is preserved despite the lack of pixel-level overlap.

\section{Experiment 3}
The previous two experiments demonstrated that we could produce a robust shape bias response in a variety of artificial neural networks using cue-conflict stimuli based on familiar categories. In this final experiment, we take one more step towards matching the developmental paradigm and measure shape bias using novel shapes and textures, differing from the categories the networks were trained on. This procedure is similar to that taken by \citeA{ritter2017cognitive} in which they also measured shape bias in DNNs using triplets of novel stimuli. However, while their work focused on testing shape versus color bias, our aim is to test shape versus texture bias. 

\textbf{Dataset.} We created 256 unique novel stimuli using 16 high-quality textures \cite{brodatz1966textures} and 16 simple shapes \cite{parks2020statistical} by overlaying random patches of each texture with a mask of each shape, resulting in misaligned textures between texture matches. As in Experiment 2, we varied the size and alignment of these novel stimuli, and generated all possible triplets as described in the \textbf{Measuring Shape Bias} section. Unlike Experiments 1 and 2, all anchor stimulus classes were equally represented in these datasets, so shape bias is measured once for a given condition using all $57,600$ possible triplets. See Figure~\ref{fig:procedure}(a) and the bottom row of Figure~\ref{fig:stimuli} for examples of these stimuli. 

\textbf{Results and Discussion.}
When the novel stimuli are aligned in shape, all models were shape biased for the smallest size (see Figure~\ref{fig:align}, bottom-left). As shape size increased, most models showed a decrease in shape bias (whether aligned or not), leading some models to display a texture bias in the largest size conditions. To our surprise, DINO ResNet-50 was an  exception, showing an increase in shape bias as the shapes become larger. It is not entirely clear why DINO behaves qualitatively differently to the other models we tested, and we leave this as an open question for future research.

When the stimuli were unaligned in shape, all models made fewer shape-based decisions when compared to the aligned stimuli (see Figure~\ref{fig:align}, bottom-right); the one exception was ResNet-50, which maintained the same pattern of behavior regardless of shape alignment. This drop may be due to the impoverished nature of the silhouettes compared to natural images, or the fact these stimuli are further out-of-distribution compared to pre-training. An advantage of this test setting is the neutral performance of the untrained random ResNet-50, which hovers around chance. Notably, ResNet-50, ViT-B/16, and CLIP ViT-B/16 all exhibited a shape bias relative to their untrained counterparts, indicating that they acquired more sensitivity to shape for novel stimuli during their training. This is consistent with the developmental picture, in which children acquire a shape bias as they learn more words \cite{smith2002object, gershkoff2004shape}.  

\section{Discussion}
We outlined a more developmentally consistent procedure for testing shape bias in artificial neural networks. One advantage of using embedding similarity instead of model output is that it allows for the measurement of shape bias for both pre-trained and randomly-initialized models, and for both familiar and novel stimulus categories. We measured shape bias using this procedure and three different types of cue-conflict test stimuli for a range of architectures with varying pre-training data and learning objectives. In Experiment 1, we augmented the cue-conflict stimuli used by \citeA{geirhos2019imagenet} by removing the textured background in varying degrees. We found that shape bias in all models increased as background texture salience decreased, demonstrating that the sensitivity of shape bias measurements could be explained due to characteristics of the cue-conflict stimuli. In Experiment 2, we tested whether the shape-based responses in Experiment 1 are robust to variations in shape size and position. We found that all models showed more shape-based than texture-based responses regardless of stimulus size or alignment. Finally, in Experiment 3, we generated triplets of stimuli with novel shapes and textures that varied in size and position over a white background. We found that the ImageNet pre-trained ResNet-50 and ViT-B/16 models exhibited a marked increase in shape bias over their randomly initialized variants, especially when the stimuli were randomly positioned, suggesting that these models had learned an inductive bias for shape that generalized to novel stimuli. Furthermore, shape bias across models was generally lower for the novel stimuli compared to the adapted stimuli containing familiar categories in Experiment 2.  

Our results suggest that the previously observed texture bias in DNNs arises, in part, from how texture is emphasized in both the foreground and background \cite{geirhos2019imagenet}. In fact, our results from Experiments 1 and 2 show that, despite variability in architecture and training, most models demonstrate a robust shape bias given stimuli with clean backgrounds when tested using a procedure that more closely aligns with developmental tests. However, there are still discrepancies between human and machine behavior that deserve more exploration.  \citeA{geirhos2019imagenet} observed that humans tested using the original cue-conflict stimuli still showed a strong shape bias despite the exaggerated texture present in the stimuli. One possible explanation is that humans' shape decisions were influenced by the specific testing procedure used. In their experiments, despite the neutral instructions to not bias judgments towards shape or texture, participants selected their classification decisions by clicking on one of 16 shape-based icons \cite{geirhos2018generalisation}. It is possible that the participants identified both the shape and texture classes in many stimuli, and since the task was somewhat ambiguous, the icons could have encouraged them to respond based on shape. Alternatively, models may be responding differently than people due to difficulties in segmenting the foreground of the images (i.e. the shape) from the background. Further work is needed to understand precisely why neural networks struggle with exaggerated background textures and if there are other ways for DNNs to match human behavior on the original cue-conflict stimuli.   

It is worth noting that both \citeA{baker2018deep} and \citeA{geirhos2019imagenet} performed other experiments in which they fill a silhouette (shape) from one class with a texture from another, similar to our $\alpha = 1$ condition in Experiment 1. Unlike our results, they find that texture is emphasized over shape. Their stimuli, however, are notably different. Filling a silhouette with a cutout from another image can not only erase internal features of the first category but can also introduce recognizable features from a different familiar category. These differences further highlight that the degree of shape bias in current models is highly dependent on task and stimulus details. Of the evaluations we consider, our Experiment 3 most closely mirrors the developmental procedure as run with children, although no single test seems to tell the whole story.

The developmental account of the interrelationship between the shape bias and word learning in children seems to align with our observations that models with pre-training on labels or linguistic data exhibit a stronger shape bias than their untrained counterparts. However, the strong shape preference exhibited by the self-supervised models in some conditions is somewhat puzzling from the developmental point of view given their complete lack of language exposure. It remains unclear why a robust shape bias is observed in all models regardless of their pre-training objective. Moreover, it is unclear why models display substantially more variation than people with regards to the shape bias.

As deep neural networks have continued to demonstrate impressive performance on visual tasks and have thus seen an expanding range of applications, it has become increasingly important to understand the nature of the biases they employ to classify objects and how these biases compare to those learned by humans. The recent explorations into the existence of a shape bias in DNNs have yielded a number of unexpected and significant results. In this paper, we add to this line of research by measuring shape versus texture bias in a range of DNN architectures using a procedure that more closely matches the developmental setup, highlighting that close attention needs to be paid to all aspects of the evaluation in order to conduct a ``species-fair comparison'' \cite{firestone2020performance}. Ultimately, the shape bias is only one of many inductive biases employed by the human mind to understand and make sense of the world. Both machine learning and cognitive science stand to gain from further investigation into models that capture a range of inductive biases as well as methods for most fruitfully comparing human and machine intelligence. 

% WK: here's some potential ideas for discussion paragraphs
% - How do our results impact the claim that CNNs demonstrate a texture bias? Suggests the finding may have been an artifact of the cue conflict stimuli, results from Exp 1 and 2 suggest a robust preference for shape match rather than texture, using the same stimuli but matching the developmental procedure
% - Why were humans shape biased in the Geirhos experiments? Could mention the shape-based icons of the task biasing participants towards shape responses, something that wasn’t available to models (I think there's other examples of small experimental manipulations that influence shape bias, will do some reading to find relevant papers to cite!)
% - Why do we see evidence of shape bias in all models, regardless if they are supervised, self-supervised, contrastive? Moreover, degree of shape bias across models are not consistent across experiments, still open questions, esp. trying to compare these back to human shape bias results!

\section{Acknowledgments}
We thank Erin Grant, Pat Little, Pam Osborn Popp, and the anonymous reviewers for their valuable feedback. This work was supported by NIH grant R90DA043849.

\bibliographystyle{apacite}

\setlength{\bibleftmargin}{.125in}
\setlength{\bibindent}{-\bibleftmargin}

\bibliography{CogSci_Template}

\begin{thebibliography}{}

\bibitem [\protect \citeauthoryear {%
Baker%
, Lu%
, Erlikhman%
\BCBL {}\ \BBA {} Kellman%
}{%
Baker%
\ \protect \BOthers {.}}{%
{\protect \APACyear {2018}}%
}]{%
baker2018deep}
\APACinsertmetastar {%
baker2018deep}%
\begin{APACrefauthors}%
Baker, N.%
, Lu, H.%
, Erlikhman, G.%
\BCBL {}\ \BBA {} Kellman, P\BPBI J.%
\end{APACrefauthors}%
\unskip\
\newblock
\APACrefYearMonthDay{2018}{}{}.
\newblock
{\BBOQ}\APACrefatitle {Deep convolutional networks do not classify based on
  global object shape} {Deep convolutional networks do not classify based on
  global object shape}.{\BBCQ}
\newblock
\APACjournalVolNumPages{PLoS Computational Biology}{14}{12}{e1006613}.
\PrintBackRefs{\CurrentBib}

\bibitem [\protect \citeauthoryear {%
Biederman%
}{%
Biederman%
}{%
{\protect \APACyear {1995}}%
}]{%
biederman1995visual}
\APACinsertmetastar {%
biederman1995visual}%
\begin{APACrefauthors}%
Biederman, I.%
\end{APACrefauthors}%
\unskip\
\newblock
\APACrefYearMonthDay{1995}{}{}.
\newblock
{\BBOQ}\APACrefatitle {Visual object recognition} {Visual object
  recognition}.{\BBCQ}
\newblock
\BIn{} S\BPBI M.~Kosslyn\ \BBA {} D\BPBI N.~Osherson\ (\BEDS), \APACrefbtitle
  {Visual cognition: An invitation to cognitive science} {Visual cognition: An
  invitation to cognitive science}\ (\BVOL~2, \BPG~121–165).
\newblock
\APACaddressPublisher{}{MIT press Cambridge, MA, USA}.
\PrintBackRefs{\CurrentBib}

\bibitem [\protect \citeauthoryear {%
Brendel%
\ \BBA {} Bethge%
}{%
Brendel%
\ \BBA {} Bethge%
}{%
{\protect \APACyear {2019}}%
}]{%
brendel2019approximating}
\APACinsertmetastar {%
brendel2019approximating}%
\begin{APACrefauthors}%
Brendel, W.%
\BCBT {}\ \BBA {} Bethge, M.%
\end{APACrefauthors}%
\unskip\
\newblock
\APACrefYearMonthDay{2019}{}{}.
\newblock
{\BBOQ}\APACrefatitle {{Approximating CNNs with bag-of-local-features models
  works surprisingly well on ImageNet}} {{Approximating CNNs with
  bag-of-local-features models works surprisingly well on ImageNet}}.{\BBCQ}
\newblock
\APACjournalVolNumPages{arXiv preprint arXiv:1904.00760}{}{}{}.
\PrintBackRefs{\CurrentBib}

\bibitem [\protect \citeauthoryear {%
Brodatz%
}{%
Brodatz%
}{%
{\protect \APACyear {1966}}%
}]{%
brodatz1966textures}
\APACinsertmetastar {%
brodatz1966textures}%
\begin{APACrefauthors}%
Brodatz, P.%
\end{APACrefauthors}%
\unskip\
\newblock
\APACrefYear{1966}.
\newblock
\APACrefbtitle {Textures: a photographic album for artists and designers}
  {Textures: a photographic album for artists and designers}.
\newblock
\APACaddressPublisher{}{Dover publications}.
\PrintBackRefs{\CurrentBib}

\bibitem [\protect \citeauthoryear {%
Caron%
\ \protect \BOthers {.}}{%
Caron%
\ \protect \BOthers {.}}{%
{\protect \APACyear {2021}}%
}]{%
caron2021emerging}
\APACinsertmetastar {%
caron2021emerging}%
\begin{APACrefauthors}%
Caron, M.%
, Touvron, H.%
, Misra, I.%
, J{\'e}gou, H.%
, Mairal, J.%
, Bojanowski, P.%
\BCBL {}\ \BBA {} Joulin, A.%
\end{APACrefauthors}%
\unskip\
\newblock
\APACrefYearMonthDay{2021}{}{}.
\newblock
{\BBOQ}\APACrefatitle {Emerging properties in self-supervised vision
  transformers} {Emerging properties in self-supervised vision
  transformers}.{\BBCQ}
\newblock
\APACjournalVolNumPages{arXiv preprint arXiv:2104.14294}{}{}{}.
\PrintBackRefs{\CurrentBib}

\bibitem [\protect \citeauthoryear {%
Colunga%
\ \BBA {} Smith%
}{%
Colunga%
\ \BBA {} Smith%
}{%
{\protect \APACyear {2005}}%
}]{%
colunga2005lexicon}
\APACinsertmetastar {%
colunga2005lexicon}%
\begin{APACrefauthors}%
Colunga, E.%
\BCBT {}\ \BBA {} Smith, L\BPBI B.%
\end{APACrefauthors}%
\unskip\
\newblock
\APACrefYearMonthDay{2005}{}{}.
\newblock
{\BBOQ}\APACrefatitle {From the lexicon to expectations about kinds: A role for
  associative learning.} {From the lexicon to expectations about kinds: A role
  for associative learning.}{\BBCQ}
\newblock
\APACjournalVolNumPages{Psychological Review}{112}{2}{347}.
\PrintBackRefs{\CurrentBib}

\bibitem [\protect \citeauthoryear {%
Deng%
\ \protect \BOthers {.}}{%
Deng%
\ \protect \BOthers {.}}{%
{\protect \APACyear {2009}}%
}]{%
deng2009imagenet}
\APACinsertmetastar {%
deng2009imagenet}%
\begin{APACrefauthors}%
Deng, J.%
, Dong, W.%
, Socher, R.%
, Li, L\BHBI J.%
, Li, K.%
\BCBL {}\ \BBA {} Fei-Fei, L.%
\end{APACrefauthors}%
\unskip\
\newblock
\APACrefYearMonthDay{2009}{}{}.
\newblock
{\BBOQ}\APACrefatitle {ImageNet: A large-scale hierarchical image database}
  {Imagenet: A large-scale hierarchical image database}.{\BBCQ}
\newblock
\BIn{} \APACrefbtitle {{IEEE conference on Computer Vision and Pattern
  Recognition}} {{IEEE conference on Computer Vision and Pattern Recognition}}\
  (\BPGS\ 248--255).
\PrintBackRefs{\CurrentBib}

\bibitem [\protect \citeauthoryear {%
Diesendruck%
\ \BBA {} Bloom%
}{%
Diesendruck%
\ \BBA {} Bloom%
}{%
{\protect \APACyear {2003}}%
}]{%
diesendruck2003specific}
\APACinsertmetastar {%
diesendruck2003specific}%
\begin{APACrefauthors}%
Diesendruck, G.%
\BCBT {}\ \BBA {} Bloom, P.%
\end{APACrefauthors}%
\unskip\
\newblock
\APACrefYearMonthDay{2003}{}{}.
\newblock
{\BBOQ}\APACrefatitle {How specific is the shape bias?} {How specific is the
  shape bias?}{\BBCQ}
\newblock
\APACjournalVolNumPages{Child Development}{74}{1}{168--178}.
\PrintBackRefs{\CurrentBib}

\bibitem [\protect \citeauthoryear {%
Dosovitskiy%
\ \protect \BOthers {.}}{%
Dosovitskiy%
\ \protect \BOthers {.}}{%
{\protect \APACyear {2021}}%
}]{%
dosovitskiy2021image}
\APACinsertmetastar {%
dosovitskiy2021image}%
\begin{APACrefauthors}%
Dosovitskiy, A.%
, Beyer, L.%
, Kolesnikov, A.%
, Weissenborn, D.%
, Zhai, X.%
, Unterthiner, T.%
\BDBL {}others%
\end{APACrefauthors}%
\unskip\
\newblock
\APACrefYearMonthDay{2021}{}{}.
\newblock
{\BBOQ}\APACrefatitle {An image is worth 16x16 words: Transformers for image
  recognition at scale} {An image is worth 16x16 words: Transformers for image
  recognition at scale}.{\BBCQ}
\newblock
\BIn{} \APACrefbtitle {{International Conference on Learning Representations}.}
  {{International Conference on Learning Representations}.}
\PrintBackRefs{\CurrentBib}

\bibitem [\protect \citeauthoryear {%
Feinman%
\ \BBA {} Lake%
}{%
Feinman%
\ \BBA {} Lake%
}{%
{\protect \APACyear {2018}}%
}]{%
feinman2018learning}
\APACinsertmetastar {%
feinman2018learning}%
\begin{APACrefauthors}%
Feinman, R.%
\BCBT {}\ \BBA {} Lake, B\BPBI M.%
\end{APACrefauthors}%
\unskip\
\newblock
\APACrefYearMonthDay{2018}{}{}.
\newblock
{\BBOQ}\APACrefatitle {Learning inductive biases with simple neural networks}
  {Learning inductive biases with simple neural networks}.{\BBCQ}
\newblock
\BIn{} \APACrefbtitle {{Proceedings of the 40th Annual Conference of the
  Cognitive Science Society}.} {{Proceedings of the 40th Annual Conference of
  the Cognitive Science Society}.}
\PrintBackRefs{\CurrentBib}

\bibitem [\protect \citeauthoryear {%
Firestone%
}{%
Firestone%
}{%
{\protect \APACyear {2020}}%
}]{%
firestone2020performance}
\APACinsertmetastar {%
firestone2020performance}%
\begin{APACrefauthors}%
Firestone, C.%
\end{APACrefauthors}%
\unskip\
\newblock
\APACrefYearMonthDay{2020}{}{}.
\newblock
{\BBOQ}\APACrefatitle {Performance vs. competence in human--machine
  comparisons} {Performance vs. competence in human--machine
  comparisons}.{\BBCQ}
\newblock
\APACjournalVolNumPages{Proceedings of the National Academy of
  Sciences}{117}{43}{26562--26571}.
\PrintBackRefs{\CurrentBib}

\bibitem [\protect \citeauthoryear {%
Gatys%
, Ecker%
\BCBL {}\ \BBA {} Bethge%
}{%
Gatys%
\ \protect \BOthers {.}}{%
{\protect \APACyear {2016}}%
}]{%
gatys2016image}
\APACinsertmetastar {%
gatys2016image}%
\begin{APACrefauthors}%
Gatys, L\BPBI A.%
, Ecker, A\BPBI S.%
\BCBL {}\ \BBA {} Bethge, M.%
\end{APACrefauthors}%
\unskip\
\newblock
\APACrefYearMonthDay{2016}{}{}.
\newblock
{\BBOQ}\APACrefatitle {Image style transfer using convolutional neural
  networks} {Image style transfer using convolutional neural networks}.{\BBCQ}
\newblock
\BIn{} \APACrefbtitle {Proceedings of the IEEE conference on computer vision
  and pattern recognition} {Proceedings of the ieee conference on computer
  vision and pattern recognition}\ (\BPGS\ 2414--2423).
\PrintBackRefs{\CurrentBib}

\bibitem [\protect \citeauthoryear {%
Geirhos%
\ \protect \BOthers {.}}{%
Geirhos%
\ \protect \BOthers {.}}{%
{\protect \APACyear {2021}}%
}]{%
geirhos2021partial}
\APACinsertmetastar {%
geirhos2021partial}%
\begin{APACrefauthors}%
Geirhos, R.%
, Narayanappa, K.%
, Mitzkus, B.%
, Thieringer, T.%
, Bethge, M.%
, Wichmann, F\BPBI A.%
\BCBL {}\ \BBA {} Brendel, W.%
\end{APACrefauthors}%
\unskip\
\newblock
\APACrefYearMonthDay{2021}{}{}.
\newblock
{\BBOQ}\APACrefatitle {Partial success in closing the gap between human and
  machine vision} {Partial success in closing the gap between human and machine
  vision}.{\BBCQ}
\newblock
\BIn{} \APACrefbtitle {{Advances in Neural Information Processing Systems}.}
  {{Advances in Neural Information Processing Systems}.}
\PrintBackRefs{\CurrentBib}

\bibitem [\protect \citeauthoryear {%
Geirhos%
\ \protect \BOthers {.}}{%
Geirhos%
\ \protect \BOthers {.}}{%
{\protect \APACyear {2019}}%
}]{%
geirhos2019imagenet}
\APACinsertmetastar {%
geirhos2019imagenet}%
\begin{APACrefauthors}%
Geirhos, R.%
, Rubisch, P.%
, Michaelis, C.%
, Bethge, M.%
, Wichmann, F\BPBI A.%
\BCBL {}\ \BBA {} Brendel, W.%
\end{APACrefauthors}%
\unskip\
\newblock
\APACrefYearMonthDay{2019}{}{}.
\newblock
{\BBOQ}\APACrefatitle {{ImageNet-trained CNNs are biased towards texture;
  increasing shape bias improves accuracy and robustness}} {{ImageNet-trained
  CNNs are biased towards texture; increasing shape bias improves accuracy and
  robustness}}.{\BBCQ}
\newblock
\BIn{} \APACrefbtitle {International Conference on Learning Representations.}
  {International conference on learning representations.}
\PrintBackRefs{\CurrentBib}

\bibitem [\protect \citeauthoryear {%
Geirhos%
\ \protect \BOthers {.}}{%
Geirhos%
\ \protect \BOthers {.}}{%
{\protect \APACyear {2018}}%
}]{%
geirhos2018generalisation}
\APACinsertmetastar {%
geirhos2018generalisation}%
\begin{APACrefauthors}%
Geirhos, R.%
, Temme, C\BPBI R\BPBI M.%
, Rauber, J.%
, Sch{\"u}tt, H\BPBI H.%
, Bethge, M.%
\BCBL {}\ \BBA {} Wichmann, F\BPBI A.%
\end{APACrefauthors}%
\unskip\
\newblock
\APACrefYearMonthDay{2018}{}{}.
\newblock
{\BBOQ}\APACrefatitle {Generalisation in humans and deep neural networks}
  {Generalisation in humans and deep neural networks}.{\BBCQ}
\newblock
\BIn{} \APACrefbtitle {{Advances in Neural Information Processing Systems}}
  {{Advances in Neural Information Processing Systems}}\ (\BPGS\ 7549--7561).
\PrintBackRefs{\CurrentBib}

\bibitem [\protect \citeauthoryear {%
Gershkoff-Stowe%
\ \BBA {} Smith%
}{%
Gershkoff-Stowe%
\ \BBA {} Smith%
}{%
{\protect \APACyear {2004}}%
}]{%
gershkoff2004shape}
\APACinsertmetastar {%
gershkoff2004shape}%
\begin{APACrefauthors}%
Gershkoff-Stowe, L.%
\BCBT {}\ \BBA {} Smith, L\BPBI B.%
\end{APACrefauthors}%
\unskip\
\newblock
\APACrefYearMonthDay{2004}{}{}.
\newblock
{\BBOQ}\APACrefatitle {Shape and the first hundred nouns} {Shape and the first
  hundred nouns}.{\BBCQ}
\newblock
\APACjournalVolNumPages{Child development}{75}{4}{1098--1114}.
\PrintBackRefs{\CurrentBib}

\bibitem [\protect \citeauthoryear {%
He%
, Zhang%
, Ren%
\BCBL {}\ \BBA {} Sun%
}{%
He%
\ \protect \BOthers {.}}{%
{\protect \APACyear {2016}}%
}]{%
he2016deep}
\APACinsertmetastar {%
he2016deep}%
\begin{APACrefauthors}%
He, K.%
, Zhang, X.%
, Ren, S.%
\BCBL {}\ \BBA {} Sun, J.%
\end{APACrefauthors}%
\unskip\
\newblock
\APACrefYearMonthDay{2016}{}{}.
\newblock
{\BBOQ}\APACrefatitle {Deep residual learning for image recognition} {Deep
  residual learning for image recognition}.{\BBCQ}
\newblock
\BIn{} \APACrefbtitle {{Proceedings of the IEEE conference on Computer Vision
  and Pattern Recognition}} {{Proceedings of the IEEE conference on Computer
  Vision and Pattern Recognition}}\ (\BPGS\ 770--778).
\PrintBackRefs{\CurrentBib}

\bibitem [\protect \citeauthoryear {%
Hermann%
, Chen%
\BCBL {}\ \BBA {} Kornblith%
}{%
Hermann%
\ \protect \BOthers {.}}{%
{\protect \APACyear {2020}}%
}]{%
hermann2020origins}
\APACinsertmetastar {%
hermann2020origins}%
\begin{APACrefauthors}%
Hermann, K.%
, Chen, T.%
\BCBL {}\ \BBA {} Kornblith, S.%
\end{APACrefauthors}%
\unskip\
\newblock
\APACrefYearMonthDay{2020}{}{}.
\newblock
{\BBOQ}\APACrefatitle {The Origins and Prevalence of Texture Bias in
  Convolutional Neural Networks} {The origins and prevalence of texture bias in
  convolutional neural networks}.{\BBCQ}
\newblock
\APACjournalVolNumPages{Advances in Neural Information Processing
  Systems}{33}{}{}.
\PrintBackRefs{\CurrentBib}

\bibitem [\protect \citeauthoryear {%
Kemp%
, Perfors%
\BCBL {}\ \BBA {} Tenenbaum%
}{%
Kemp%
\ \protect \BOthers {.}}{%
{\protect \APACyear {2007}}%
}]{%
kemp2007learning}
\APACinsertmetastar {%
kemp2007learning}%
\begin{APACrefauthors}%
Kemp, C.%
, Perfors, A.%
\BCBL {}\ \BBA {} Tenenbaum, J\BPBI B.%
\end{APACrefauthors}%
\unskip\
\newblock
\APACrefYearMonthDay{2007}{}{}.
\newblock
{\BBOQ}\APACrefatitle {{Learning overhypotheses with hierarchical Bayesian
  models}} {{Learning overhypotheses with hierarchical Bayesian
  models}}.{\BBCQ}
\newblock
\APACjournalVolNumPages{Developmental Science}{10}{3}{307--321}.
\PrintBackRefs{\CurrentBib}

\bibitem [\protect \citeauthoryear {%
Krizhevsky%
, Sutskever%
\BCBL {}\ \BBA {} Hinton%
}{%
Krizhevsky%
\ \protect \BOthers {.}}{%
{\protect \APACyear {2012}}%
}]{%
krizhevsky2012imagenet}
\APACinsertmetastar {%
krizhevsky2012imagenet}%
\begin{APACrefauthors}%
Krizhevsky, A.%
, Sutskever, I.%
\BCBL {}\ \BBA {} Hinton, G\BPBI E.%
\end{APACrefauthors}%
\unskip\
\newblock
\APACrefYearMonthDay{2012}{}{}.
\newblock
{\BBOQ}\APACrefatitle {Imagenet classification with deep convolutional neural
  networks} {Imagenet classification with deep convolutional neural
  networks}.{\BBCQ}
\newblock
\BIn{} \APACrefbtitle {{Advances in Neural Information Processing Systems}}
  {{Advances in Neural Information Processing Systems}}\ (\BVOL~25, \BPGS\
  1097--1105).
\PrintBackRefs{\CurrentBib}

\bibitem [\protect \citeauthoryear {%
Landau%
, Smith%
\BCBL {}\ \BBA {} Jones%
}{%
Landau%
\ \protect \BOthers {.}}{%
{\protect \APACyear {1988}}%
}]{%
landau1988importance}
\APACinsertmetastar {%
landau1988importance}%
\begin{APACrefauthors}%
Landau, B.%
, Smith, L\BPBI B.%
\BCBL {}\ \BBA {} Jones, S\BPBI S.%
\end{APACrefauthors}%
\unskip\
\newblock
\APACrefYearMonthDay{1988}{}{}.
\newblock
{\BBOQ}\APACrefatitle {The importance of shape in early lexical learning} {The
  importance of shape in early lexical learning}.{\BBCQ}
\newblock
\APACjournalVolNumPages{Cognitive Development}{3}{3}{299--321}.
\PrintBackRefs{\CurrentBib}

\bibitem [\protect \citeauthoryear {%
Orhan%
, Gupta%
\BCBL {}\ \BBA {} Lake%
}{%
Orhan%
\ \protect \BOthers {.}}{%
{\protect \APACyear {2020}}%
}]{%
orhan2020self}
\APACinsertmetastar {%
orhan2020self}%
\begin{APACrefauthors}%
Orhan, E.%
, Gupta, V.%
\BCBL {}\ \BBA {} Lake, B\BPBI M.%
\end{APACrefauthors}%
\unskip\
\newblock
\APACrefYearMonthDay{2020}{}{}.
\newblock
{\BBOQ}\APACrefatitle {Self-supervised learning through the eyes of a child}
  {Self-supervised learning through the eyes of a child}.{\BBCQ}
\newblock
\BIn{} \APACrefbtitle {{Advances in Neural Information Processing Systems}}
  {{Advances in Neural Information Processing Systems}}\ (\BVOL~33).
\PrintBackRefs{\CurrentBib}

\bibitem [\protect \citeauthoryear {%
Parks%
, Griffith%
, Armstrong%
\BCBL {}\ \BBA {} Stevenson%
}{%
Parks%
\ \protect \BOthers {.}}{%
{\protect \APACyear {2020}}%
}]{%
parks2020statistical}
\APACinsertmetastar {%
parks2020statistical}%
\begin{APACrefauthors}%
Parks, K\BPBI M.%
, Griffith, L\BPBI A.%
, Armstrong, N\BPBI B.%
\BCBL {}\ \BBA {} Stevenson, R\BPBI A.%
\end{APACrefauthors}%
\unskip\
\newblock
\APACrefYearMonthDay{2020}{}{}.
\newblock
{\BBOQ}\APACrefatitle {Statistical Learning and Social Competency: The
  Mediating Role of Language} {Statistical learning and social competency: The
  mediating role of language}.{\BBCQ}
\newblock
\APACjournalVolNumPages{Scientific Reports}{10}{1}{1--15}.
\PrintBackRefs{\CurrentBib}

\bibitem [\protect \citeauthoryear {%
Radford%
\ \protect \BOthers {.}}{%
Radford%
\ \protect \BOthers {.}}{%
{\protect \APACyear {2021}}%
}]{%
radford2021learning}
\APACinsertmetastar {%
radford2021learning}%
\begin{APACrefauthors}%
Radford, A.%
, Kim, J\BPBI W.%
, Hallacy, C.%
, Ramesh, A.%
, Goh, G.%
, Agarwal, S.%
\BDBL {}others%
\end{APACrefauthors}%
\unskip\
\newblock
\APACrefYearMonthDay{2021}{}{}.
\newblock
{\BBOQ}\APACrefatitle {Learning transferable visual models from natural
  language supervision} {Learning transferable visual models from natural
  language supervision}.{\BBCQ}
\newblock
\APACjournalVolNumPages{arXiv preprint arXiv:2103.00020}{}{}{}.
\PrintBackRefs{\CurrentBib}

\bibitem [\protect \citeauthoryear {%
Ritter%
, Barrett%
, Santoro%
\BCBL {}\ \BBA {} Botvinick%
}{%
Ritter%
\ \protect \BOthers {.}}{%
{\protect \APACyear {2017}}%
}]{%
ritter2017cognitive}
\APACinsertmetastar {%
ritter2017cognitive}%
\begin{APACrefauthors}%
Ritter, S.%
, Barrett, D\BPBI G.%
, Santoro, A.%
\BCBL {}\ \BBA {} Botvinick, M\BPBI M.%
\end{APACrefauthors}%
\unskip\
\newblock
\APACrefYearMonthDay{2017}{}{}.
\newblock
{\BBOQ}\APACrefatitle {Cognitive psychology for deep neural networks: A shape
  bias case study} {Cognitive psychology for deep neural networks: A shape bias
  case study}.{\BBCQ}
\newblock
\BIn{} \APACrefbtitle {{International Conference on Machine Learning}}
  {{International Conference on Machine Learning}}\ (\BPGS\ 2940--2949).
\PrintBackRefs{\CurrentBib}

\bibitem [\protect \citeauthoryear {%
Samuelson%
}{%
Samuelson%
}{%
{\protect \APACyear {2002}}%
}]{%
samuelson2002statistical}
\APACinsertmetastar {%
samuelson2002statistical}%
\begin{APACrefauthors}%
Samuelson, L\BPBI K.%
\end{APACrefauthors}%
\unskip\
\newblock
\APACrefYearMonthDay{2002}{}{}.
\newblock
{\BBOQ}\APACrefatitle {Statistical regularities in vocabulary guide language
  acquisition in connectionist models and 15-20-month-olds.} {Statistical
  regularities in vocabulary guide language acquisition in connectionist models
  and 15-20-month-olds.}{\BBCQ}
\newblock
\APACjournalVolNumPages{Developmental Psychology}{38}{6}{1016}.
\PrintBackRefs{\CurrentBib}

\bibitem [\protect \citeauthoryear {%
Smith%
, Jones%
, Landau%
, Gershkoff-Stowe%
\BCBL {}\ \BBA {} Samuelson%
}{%
Smith%
\ \protect \BOthers {.}}{%
{\protect \APACyear {2002}}%
}]{%
smith2002object}
\APACinsertmetastar {%
smith2002object}%
\begin{APACrefauthors}%
Smith, L\BPBI B.%
, Jones, S\BPBI S.%
, Landau, B.%
, Gershkoff-Stowe, L.%
\BCBL {}\ \BBA {} Samuelson, L.%
\end{APACrefauthors}%
\unskip\
\newblock
\APACrefYearMonthDay{2002}{}{}.
\newblock
{\BBOQ}\APACrefatitle {Object name learning provides on-the-job training for
  attention} {Object name learning provides on-the-job training for
  attention}.{\BBCQ}
\newblock
\APACjournalVolNumPages{Psychological Science}{13}{1}{13--19}.
\PrintBackRefs{\CurrentBib}

\bibitem [\protect \citeauthoryear {%
Tuli%
, Dasgupta%
, Grant%
\BCBL {}\ \BBA {} Griffiths%
}{%
Tuli%
\ \protect \BOthers {.}}{%
{\protect \APACyear {2021}}%
}]{%
tuli2021convolutional}
\APACinsertmetastar {%
tuli2021convolutional}%
\begin{APACrefauthors}%
Tuli, S.%
, Dasgupta, I.%
, Grant, E.%
\BCBL {}\ \BBA {} Griffiths, T\BPBI L.%
\end{APACrefauthors}%
\unskip\
\newblock
\APACrefYearMonthDay{2021}{}{}.
\newblock
{\BBOQ}\APACrefatitle {Are Convolutional Neural Networks or Transformers more
  like human vision?} {Are convolutional neural networks or transformers more
  like human vision?}{\BBCQ}
\newblock
\BIn{} \APACrefbtitle {{Proceedings of the 43rd Annual Conference of the
  Cognitive Science Society}.} {{Proceedings of the 43rd Annual Conference of
  the Cognitive Science Society}.}
\PrintBackRefs{\CurrentBib}

\bibitem [\protect \citeauthoryear {%
Xie%
, Girshick%
, Doll{\'a}r%
, Tu%
\BCBL {}\ \BBA {} He%
}{%
Xie%
\ \protect \BOthers {.}}{%
{\protect \APACyear {2017}}%
}]{%
xie2017aggregated}
\APACinsertmetastar {%
xie2017aggregated}%
\begin{APACrefauthors}%
Xie, S.%
, Girshick, R.%
, Doll{\'a}r, P.%
, Tu, Z.%
\BCBL {}\ \BBA {} He, K.%
\end{APACrefauthors}%
\unskip\
\newblock
\APACrefYearMonthDay{2017}{}{}.
\newblock
{\BBOQ}\APACrefatitle {Aggregated residual transformations for deep neural
  networks} {Aggregated residual transformations for deep neural
  networks}.{\BBCQ}
\newblock
\BIn{} \APACrefbtitle {{Proceedings of the IEEE conference on Computer Vision
  and Pattern Recognition}} {{Proceedings of the IEEE conference on Computer
  Vision and Pattern Recognition}}\ (\BPGS\ 1492--1500).
\PrintBackRefs{\CurrentBib}

\bibitem [\protect \citeauthoryear {%
Yamins%
\ \protect \BOthers {.}}{%
Yamins%
\ \protect \BOthers {.}}{%
{\protect \APACyear {2014}}%
}]{%
yamins2014performance}
\APACinsertmetastar {%
yamins2014performance}%
\begin{APACrefauthors}%
Yamins, D\BPBI L.%
, Hong, H.%
, Cadieu, C\BPBI F.%
, Solomon, E\BPBI A.%
, Seibert, D.%
\BCBL {}\ \BBA {} DiCarlo, J\BPBI J.%
\end{APACrefauthors}%
\unskip\
\newblock
\APACrefYearMonthDay{2014}{}{}.
\newblock
{\BBOQ}\APACrefatitle {Performance-optimized hierarchical models predict neural
  responses in higher visual cortex} {Performance-optimized hierarchical models
  predict neural responses in higher visual cortex}.{\BBCQ}
\newblock
\APACjournalVolNumPages{Proceedings of the National Academy of
  Sciences}{111}{23}{8619--8624}.
\PrintBackRefs{\CurrentBib}

\end{thebibliography}

\end{document}